# What can linguistics and deep learning contribute to each other?

Tal Linzen
Department of Cognitive Science
Johns Hopkins University
tal.linzen@jhu.edu
3400 N. Charles Street
Baltimore, Maryland

**Abstract***

Joe Pater's target article calls for greater interaction between neural network research and linguistics. I expand on this call and show how such interaction can benefit both fields. Linguists can contribute to research on neural networks for language technologies by clearly delineating the linguistic capabilities that can be expected of such systems, and by constructing controlled experimental paradigms that can determine whether those desiderata have been met. In the other direction, neural networks can benefit the scientific study of language by providing infrastructure for modeling human sentence processing and for evaluating the necessity of particular innate constraints on language acquisition.

*I thank Brian Leonard, Tom McCoy, Grusha Prasad and Marten van Schijndel for useful comments and discussion.

Keywords: neural networks, psycholinguistics, poverty of the stimulus, deep learning, syntax, sentence processing

In the last section of his target article, Pater (2018)[1] calls for greater fusion between linguistics and research on neural networks ("deep learning"); he focuses on ways in which linguistics can benefit from such fusion. The goal of this response is to expand on his article and demonstrate how closer interaction between deep learning and the cognitive science of language can be mutually beneficial to both fields. In one direction, the theoretical distinctions proposed by linguists provide a yardstick by which the performance of deep learning systems can be measured. As I discuss in Section 1, rather than emphasize test cases that are similar to the training examples, as is typical in machine learning research, linguistically informed evaluation focuses on cases that probe interpretable dimensions of generalization from the training set. The construction of test materials that avoid confounds in order to isolate a particular linguistic capacity requires close attention to experimental design; paradigms and methodologies from psycholinguistics will be helpful in accomplishing this goal (Section 2). In the other direction, neural networks can serve as a useful tool in the scientific study of language, by providing a computational platform for testing whether proposed innate learning constraints are necessary for language acquisition (Section 3) and for constructing models of human sentence processing (Section 4). Overall, I agree with Pater that the adoption of neural networks need not entail a rejection of generative linguistics, and vice versa; such false dichotomies between the fields will unnecessarily hinder the interactions outlined in his article and in the current response.

## 1. Linguistics as a normative standard

Systems based on neural networks have become increasingly common in technological applications over the last few years. One of the tasks that neural networks excel in consists in estimating how likely a particular word is to occur given the words that have preceded it; systems that perform this task are referred to as "language models". Neural language models, which are typically based on recurrent neural networks (RNNs), are trained using a large corpus in the following way: suppose that the network is processing the sentence *the cat is on the mat*. After the word *cat*, the network uses its existing weights to predict how likely each of the words of the language is to follow *cat*; when it is revealed that that word was in fact *is*, the network's weights are adjusted such that in the future it will assign a higher probability to *is* in a similar context. After training completes, the network is considered successful if it assigns a high probability to the words in a new sample of sentences from the same corpus.

According to this criterion, contemporary RNN language models are able to predict upcoming words much better than earlier classes of models. At the same time, high probability assigned to a randomly sampled sentence, which may well be syntactically simple, does not necessarily indicate that the model has a robust knowledge of natural language syntax. In language modeling, as well as in tasks such as machine translation, neural networks have now reached a sufficiently high level of performance that accurately measuring their syntactic abilities and improving those abilities further will require more sophisticated evaluation methods. In particular, linguistics provides us with detailed theories of the capacities that make up the knowledge of a language; these theories make it possible to identify sentences that crucially test particular linguistic abilities.

---

[1] https://drive.google.com/file/d/1c9y79m2ecwu8QCQh_69BJnNJSlpWAHJk/view

Linzen et al.'s 2016 study illustrates this strategy. Linzen et al. argued that given a pair of minimally different sentences, one of which is grammatical and one ungrammatical, a language model should assign a higher probability to the grammatical one. Their study focused on English subject-verb agreement (Elman 1991; Bock & Miller 1991). The model was presented with the first few words of the sentence leading up to the verb, and was expected to assign a higher probability to the correct form of the verb following those words. In (1), for example, the probability of *sneezes* is expected to be higher than the probability of *sneeze*:

    (1) The black cat... sneezes/sneeze?

Of course, simple cases such as (1) can be solved without sophisticated syntactic representations; for example, the model can select the form of the verb that agrees with the most recent noun, which in many cases happens to be the subject. When a sentence contains multiple noun phrases, however, determining the number of a given verb requires identifying which of those noun phrases is the verb's subject:

    (2) The **ratio** of *men* who survive to the *women* and *children* who survive... is/are?

The head of the subject, which determines the number of the verb, is marked in boldface in (2), and "attractors", or distracting nouns with the opposite number from the head, are marked in italics. Linzen et al. trained a language model on the English Wikipedia and tested it on sentences from the same corpus with subject-verb dependencies that varied in the number of attractors intervening between the head of the subject and the verb. They found that a greater number of attractors led to a higher error rate in predicting the form of the verb (for a replication and extension, see Bernardy and Lappin 2017). In Linzen et al.'s study, which used a relatively small network, the model made more than 50% errors when tested on sentences with four attractors. Later studies that used larger RNNs were able to obtain much better accuracy (Gulordava et al. 2018; Kuncoro et al. 2018; Tran et al. 2018); crucially, however, error rates in dependencies with four attractors were still about ten times as high as in dependencies with no attractors (Kuncoro et al. 2018).

An analysis of the Wikipedia corpus shows that the majority of subject-verb agreement dependencies are syntactically simple: 95% did not have any attractors, and only 0.25% had three or more attractors (Linzen et al. 2016; Kuncoro et al. 2018). This entails that the ability of the model to predict the grammatically correct form of the verb in a complex sentence will have a very small effect on the probability assigned by the model to a random sample of sentences. Evidence of the dissociation between average probability and agreement prediction accuracy was recently demonstrated by Tran et al. (2018). They compared an RNN language model to an alternative "attention-only" neural network architecture (Vaswani et al. 2017). While the attention-only model was better at predicting the average word, it made three times as many errors as the RNN when tested on dependencies with four attractors. In another study, Kuncoro et al. (2018) found that replacing the RNN with a neural network that explicitly modeled syntax reduced errors on dependencies with multiple attractors; again, success on agreement prediction in challenging sentences did not correlate with the probability assigned by the model to a randomly sampled sentence. In both of these studies, then, evaluation using specific linguistic constructions demonstrated a difference in syntactic sophistication between systems that were not clearly distinguished based on a random sample of sentences.

## 2. Psycholinguistics and experimental control

The previous section illustrated an evaluation strategy in which test sentences are selected from a corpus based on their syntactic properties. While this approach is viable for relatively simple phenomena such as subject-verb agreement, it has its limitations. It is often difficult to automatically identify sentences with a particular set of properties in a corpus. Manually parsed corpora such as the Penn Treebank are typically too small to contain a significant number of instances of challenging constructions. While automatic parsers have become quite effective in parsing most of the sentences in the corpus on which they are trained, their performance degrades when applied to other corpora and in particular to syntactically complex sentences – precisely the sentences we would like to identify. Finally, naturally occurring examples often contain confounds that complicate the interpretation of the results (see below).

As an alternative approach, controlled materials from psycholinguistic paradigms, originally designed to evaluate sensitivity to hierarchical syntactic structure in humans, can be adapted to test the syntactic abilities of neural networks. Marvin and Linzen (2018) constructed a controlled data set that evaluates a model's command of phenomena such as subject-verb agreement across object relative clauses (3) as well as constraints on the distribution of reflexive anaphora (4) and negative polarity items (5), following the human experiments of Xiang et al. (2009):

    (3) The farmer that the parents love swims/*swim.
    (4) The manager that the architects like doubted himself/*themselves.
    (5) a. No authors that the security guards like have ever been famous.
        b. *The authors that no security guards like have ever been famous.

When tested on such sentences, which would be difficult to find in a corpus, the accuracy of the RNN language model drops dramatically. Using constructed materials, then, makes it is possible to create evaluation paradigms that are more challenging and provide a clearer indication of linguistic competence than is possible with naturally occurring sentences (for a recent application of this approach to filler-gap dependencies, see Chowdhury and Zamparelli 2018).

An important concern with naturally occurring sentences is that they often contain confounds that can make the model's accuracy harder to interpret. In particular, RNNs are exceptionally prone to memorizing specific sequences of words. Consider the classic example of English question formation. To form an English question, the word that needs to be moved to the beginning of the sentence is the auxiliary of the main verb:

    (6) a. The little boy who is crying is hurt.
        b. Is the little boy who is crying hurt?
        c. *Is the little boy who crying is hurt?

Lewis and Elman (2001) showed that an RNN language model can distinguish the grammatical question (6b) from the ungrammatical one (6c). As it turns out, however, this preference can be derived even from a bigram model, which bases its word predictions only on the most recent word, and is thus clearly incapable of representing syntactic constraints. The reason for the success of the bigram model is simply that *who is* is a more frequent word sequence in English than sequences such as *who crying* (Reali & Christiansen 2005). To conclude that a system has learned a syntactic constraint, then, the materials need to be constructed in such a way that local heuristics would not be sufficient to perform the task (Kam et al. 2008).

At the very least, in order to demonstrate that the task requires syntactic sophistication, the model should be compared to a baseline model which cannot in principle represent syntax. For instance, Gulordava et al. (2018) compared their RNN language models to a model that only had access to the last four words when predicting the next word. Since they used materials from a corpus, in which dependencies are often short, even that baseline model was able to predict the correct verb with high accuracy (up to 91.5% for Russian); crucially, however, the full RNN did even better than this baseline (96.1% in Russian).

Instead of constructing controlled sets of materials ourselves, could we simply test RNNs on acceptability contrasts from linguistics articles, which often exemplify highly complex syntactic phenomena using constructed sentences? The answer is most likely negative: these contrasts may well suffer from the same confounds as naturally occurring sentences. In particular, linguists often use the simplest possible example that illustrates the grammatical phenomenon in question, and rely on the reader's knowledge of the language to infer the intended (typically infinite) class of sentences that instantiate the phenomenon. Lau et al. (2017) evaluated the ability of a number of language models to predict human acceptability judgments for sentences from a syntax textbook (Adger 2003). They found that the most effective predictor of human judgments was the lowest frequency bigram in the sentence (compare *who crying* in (6c)) — more effective, in fact, than the probability assigned to the sentence as a whole by an RNN. Successful prediction of the acceptability of sentences from a linguistics textbook, then, does not necessarily indicate that the network has developed a particular syntactic capacity; careful experimentation that controls for confounds, with a sufficient number of examples of each construction, is necessary to reach that conclusion.

Local word sequences are not the only non-syntactic cue that neural networks might exploit to make ostensibly syntactic predictions. Suppose that the model is evaluated on its prediction for (7):

(7) The dogs playing in the neighborhood park... barks/bark?

In this case, the subject of the verb can be identified based only on selectional preferences and semantic plausibility, without any structural analysis: given that neighborhoods and parks do not bark, the head of the subject must be *the dogs* (for a similar concern in evaluating sentence vector representations, see Ettinger et al. 2018). To control for this concern, Gulordava et al. (2018) constructed "colorless green ideas" sentences (Chomsky 1957), in which the content words from sentences from the corpus were replaced with random words with the same syntactic category and morphological features. A "colorless green ideas" sentence generated based on (7) might be:

    (8) The ideas rowing in the lamp economy... barks/bark?

Gulordava et al.'s RNN language models did not perform as well on such sentences as on original sentences from the corpus; at the same time, the decline in performance was not as dramatic as one might expect if the models relied only on semantic fit cues (or frequent word sequences, which are equally uninformative in "colorless green ideas" sentences). This suggests that the models did learn some of the syntactic principles underlying subject-verb agreement.

    **3. Innate biases in human language acquisition**

This section and the next one discuss two ways in which neural networks can be useful in the scientific study of language. The first area is in modeling human language acquisition. Syntactic theories often propose innate constraints on the learner's hypothesis space. These constraints are often motivated by intuitive learnability considerations: the evidence in the input is deemed to be insufficient for successful acquisition without those constraints. Yet intuitions about learnability may be unreliable; it is difficult to predict which aspects of the language might inform the acquisition of a particular phenomenon. The effectiveness of contemporary neural networks in learning from realistic corpora makes it possible to put these learnability concerns to an empirical test: First, is it indeed the case that the linguistic phenomenon in question cannot be learned from child-directed speech without the proposed constraint? Second, and equally important, does the proposed constraint in fact aid acquisition?

McCoy et al. (2018) report early results from such a project. They tested whether neural networks can learn to form English yes/no questions. Following Frank and Mathis (2007), McCoy et al. framed the problem as a transformation: the goal of the learner is to learn a mapping between a declarative sentence and a question, as in (9).

    (9) a. My walrus can giggle.
        b. Can my walrus giggle?

Example (9) is ambiguous: (9b) could have been generated from (9a) either by moving the leftmost auxiliary (the linear hypothesis), or by moving the auxiliary of the main verb (the structural hypothesis). These two rules make different predictions for complex sentences such as (10) — the structural rule predicts the correct question (11a), while the linear rule predicts the incorrect question (11b):

    (10) My walrus that will eat can giggle.
    (11) a. Can my walrus that will eat giggle?
        b. *Will my walrus that eat can giggle?

Examples such as (10) that disambiguate the two hypotheses are very infrequent, and a child may arguably never encounter them. This has motivated the hypothesis that children are innately constrained to consider only the structural hypothesis (Chomsky 1971). McCoy et al. trained a number of standard "sequence-to-sequence" RNNs to form such questions in a fragment of English, withholding crucial examples such as (10). These architectures, which serve as the basis of most neural machine translation systems, are not constrained to consider only structural hypotheses. Surprisingly, some of these architectures showed evidence of learning the structural rule: they fronted the auxiliary of the main verb when tested on questions such as (10). At the same time, the questions they generated differed in other respects from the questions that children form (Crain & Nakayama 1987). In other words, networks without an explicit structural bias performed better than would be expected if the structure sensitivity constraint were absolutely necessary (at least on this small fragment of English, which may provide more cues to structure than child directed speech), but their bias may not have been sufficiently strong.

An important challenge faced by this type of work is that the inductive biases of most neural network architectures are not well characterized. In McCoy et al.'s experiments, for example, relatively similar RNN architectures, which differed only in the details of the recurrence equation, generalized in surprisingly different ways. While some progress towards characterizing these biases can be made using careful work on simplified languages (Weiss et al. 2018), a fully interpretable characterization may not be possible.

There are, however, proposals for introducing interpretable and explicit syntactic inductive biases into neural networks. If the syntactic parse of the sentence is provided as part of the input, it can be incorporated into the computation of the output in a straightforward way: instead of processing the input from left to right, the network processes it according to the parse tree (Pollack 1990; Dyer et al. 2016). To create a representation of the sentence *the man eats pizza*, for example, a tree-shaped neural network would first create vector representations for *the man* and *eats pizza*, and then combine the two. In preliminary experiments, McCoy et al. have found that neural networks with explicit tree representations were able to learn the English question formation transformation well; they typically generated the correct output, and even when they did not, their mistakes were consistent with those reported for children. This suggests that this explicit bias, which is consistent with Chomsky's proposal, is indeed effective in learning this transformation. Of course, in modeling language acquisition we cannot assume that the parse is provided as part of the input; the learner needs to infer the appropriate hierarchical structure for each sentence. A number of methods for doing so have been proposed in the past two years. While their performance is not yet satisfactory, at least when evaluated on applied tasks (Williams et al. 2018), this is a new and rapidly evolving area; the potential of this approach for constructing neural networks that incorporate explicit linguistic biases will become clearer in the coming years.

## 4. Modeling human sentence processing

Another area in which neural networks can contribute to the scientific study of language is in modeling human sentence processing. Perhaps the most straightforward way is by providing predictability estimates (Section 1). Predictability strongly affects human reading behavior: predictable words are read more quickly and are more likely to be skipped. Even simple n-gram language models, which make their predictions based on the last n − 1 words, are effective in predicting reading behavior (Smith & Levy 2013). In fact, even though there have long been language models that base their predictions on an incremental parse of the sentence (Roark 2001; van Schijndel et al. 2013), it has proved difficult to conclusively demonstrate the benefit of such parser-based language models over simpler sequential models for modeling human predictions. This has led some to argue that syntactic information is not used during sentence comprehension (Frank & Bod 2011, though see Fossum & Levy 2012).

The limited benefit of grammar-based models for modeling human sentence processing may stem from the fact that they need to be trained on parsed corpora; such corpora are expensive to create and are typically too small to provide sufficient evidence for important syntactic patterns (e.g., verb subcategorization preferences). By contrast, RNN language models can be easily trained on larger corpora. Such models have been demonstrated to be superior to simpler n-gram models in predicting human reading behavior (Goodkind & Bicknell 2018). These improvements do not stem only from the RNN's superior ability to memorize frequent word sequences of variable length. Predictability estimates from RNN language models and grammar-based language models have been shown to be equally effective for predicting processing difficulty in garden path sentences such as (12), in which the words *was very upset* are unexpected given the initially preferred parse (van Schijndel & Linzen 2018a):

(12) Even though the girl phoned the instructor was very upset with her for missing a lesson.

Additional evidence that RNN language models implicitly track the probability of syntactic constructions that are not signaled by overt cues was given by van Schijndel and Linzen (2018b). They showed that exposure to sentences with reduced relative clauses, such as (13), makes the model more likely to predict similar structures, even though these structures are not signaled by any particular lexical item:

(13) The experienced soldiers warned about the dangers conducted the midnight raid.

Ultimately, as in the case of modeling language acquisition, the best predictors of psycholinguistic measurements may turn out to be hybrid neural networks that explicitly model the structure of the sentence (Hale et al. 2018).

Cases in which neural networks replicate a behavioral result from psycholinguistics without the theoretical machinery that is typically invoked to explain that result can be informative, as they suggest that the human behavior in question might arise from statistical patterns in the input. Equally importantly, divergences between human behavior and the behavior of a neural network can suggest ways in which the architecture can be changed to better model humans. For example, like the RNNs tested by Linzen et al. (2016) and subsequent work (see Section 1), humans also make agreement errors in the presence of a noun with the opposite number from the subject (agreement attraction errors). Linzen and Leonard (2018) compared the detailed pattern of errors made by RNNs and humans. They used the experimental materials of Bock and Cutting (1992), who compared attractors in relative clauses and prepositional phrases:

(14) a. Prepositional phrase: The demo tape from the popular rock *singers*...

    b. Relative clause: The demo tape that promoted the rock *singers*...

The same experiment also compared singular and plural subjects (*tape... singers* compared to *tapes... singer*). Replicating Bock and Cutting (1992), human participants in Linzen and Leonard's experiments made many more attraction errors when the head of the subject was singular (*tape... singers*) than when it was plural (*tapes... singer*). This number asymmetry has been attributed to markedness: plurals are argued to possess a privative feature that can "percolate" up to the singular subject and distort its representation, whereas singular nouns do not have such a feature and thus cannot interfere with the encoding of the plural subject (for a review, see Badecker & Kuminiak 2007). Perhaps surprisingly, RNNs showed a similar number asymmetry despite not having a notion of markedness. This suggests that it may be possible to explain the number asymmetry in agreement attraction based on corpus statistics (Haskell et al. 2010). At the same time, humans made slightly fewer errors when the attractor was in a relative clause, as in (14a), whereas the RNNs made many *more* errors on relative clauses than prepositional phrases. This finding suggests that there are significant differences between the syntactic representations that RNNs acquire and those of humans.

These initial results from work in psycholinguistics using RNNs raise important questions. It is certainly a pleasant surprise when a behavioral phenomenon emerges in an "off-the-shelf" RNN whose architecture does not explicitly encode that particular phenomenon, as in the case of the number asymmetry in agreement attraction. At the same time, when the behavior of network architectures available in standard software packages diverges from humans, as in the case of the asymmetry between attractors in prepositional phrases and in relative clauses, we must modify the network to accommodate the human data; precisely how to do so is often less clear than with classic models in which the empirical findings can be hand-engineered into the model.

Finally, there is a tension between the approach outlined in Section 1, in which the goal was for the network to perform without errors on a normatively defined task ("competence"), and the current section, in which the objective is to model fallible human language processing ("performance"). Clearly, it would not be possible to accomplish both goals using the same models. The development of competence and performance models may end up following two separate paths; alternatively, a performance model may be constructed from a competence model by combining it with additional assumptions reflecting noisy processing or resource-boundedness.

## 5. Conclusions

This response has outlined some of the ways in which deep learning and the scientific study of language can benefit each other. Linguists are best positioned to define the standards of linguistic competence that natural language technology should aspire to; to identify concrete examples that test whether those standards are met; and to interpret the behavior of the neural networks when they fail to meet them. While it may be possible to find naturally occurring sentences that evaluate the extent to which the system approaches these normative standards, such a corpus-based approach will most likely need to be supplemented with a controlled experimental approach, which emphasizes critical examples and minimizes the influence of potential confounds.

In the other direction, neural networks provide a useful platform for constructing models of language acquisition and online sentence processing. Much of the work in this vein has so far focused on architectures that are implemented in standard software packages. Yet there is no reason to believe that any particular architecture adopted from the engineering world will fit the needs of cognitive scientists perfectly. The models that gain popularity in applied settings are typically those that can effectively exploit very large collections of texts; these settings may not require the inductive biases that linguists have proposed to account for language acquisition from child directed speech. For example, even if neural networks with explicit syntactic structure are not adopted by the natural language technology world, such architectures are likely to be useful for testing cognitive questions. In other words, the scientific study of language will benefit not only from consuming the products of neural network research but also from actively contributing to it.

  *Conference on Computational Linguistics*. Association for Computational Linguistics, 1790–801.
FOSSUM, VICTORIA, and LEVY, ROGER. 2012. Sequential vs. hierarchical syntactic models of human incremental sentence processing. *Proceedings of the 3rd Workshop on Cognitive Modeling and Computational* Linguistics, Association for Computational Linguistics, 61–9.
FRANK, ROBERT, and MATHIS, DONALD. 2007. Transformational networks. *Proceedings of the 3rd Workshop on Psychocomputational Models of Human Language Acquisition.*
FRANK, STEFAN L., and BOD, RENS. 2011. Insensitivity of the human sentence-processing system to hierarchical structure. *Psychological Science* 22(6).829–34.
GOODKIND, ADAM, and BICKNELL, KLINTON. 2018. Predictive power of word surprisal for reading times is a linear function of language model quality. *Proceedings of the 8th Workshop on Cognitive Modeling and Computational Linguistics (CMCL 2018)*, 10–8.
GULORDAVA, KRISTINA; BOJANOWSKI, PIOTR; GRAVE, EDOUARD; LINZEN, TAL; and BARONI, MARCO. 2018. Colorless green recurrent networks dream hierarchically. *Proceedings of the 2018 Conference of the North American Chapter of the Association for Computational Linguistics: Human Language Technologies, Volume 1 (Long Papers)*, Association for Computational Linguistics, 1195–205.
HALE, JOHN; DYER, CHRIS; KUNCORO, ADHIGUNA; and BRENNAN, JONATHAN. 2018. Finding syntax in human encephalography with beam search. *Proceedings of the 56th Annual Meeting of the Association for Computational Linguistics (Volume 1: Long Papers),* Association for Computational Linguistics, 2727–36.
HASKELL, TODD R.; THORNTON, ROBERT; and MACDONALD, MARYELLEN C. 2010. Experience and grammatical agreement: Statistical learning shapes number agreement production. *Cognition*, 114(2).151–64.
KAM, XUÂN-NGA C.; STOYNESHKA, IGLIKA; TORNYOVA, LIDIYA; FODOR, JANET D.; and SAKAS, WILLIAM G. 2008. Bigrams and the richness of the stimulus. *Cognitive Science*, 32(4).771–87.
KUNCORO, ADHIGUNA; DYER, CHRIS; HALE, JOHN; YOGATAMA, DANI; CLARK, STEPHEN; and BLUNSOM, PHIL. 2018. LSTMs can learn syntax-sensitive dependencies well, but modeling structure makes them better. *Proceedings of the 56th Annual Meeting of the Association for Computational Linguistics (Volume 1: Long Papers)*, Association for Computational Linguistics, 1426–36.
LAU, JEY HAN; CLARK, ALEXANDER; & LAPPIN, SHALOM. 2017. Grammaticality, acceptability, and probability: A probabilistic view of linguistic knowledge. *Cognitive Science* 41(5).1202–47.
LEWIS, JOHN D., & ELMAN, JEFFREY L. 2001. Learnability and the statistical structure of language: Poverty of stimulus arguments revisited. *Proceedings of the 26th Annual Boston University Conference on Language Development*, 359–70.
LINZEN, TAL; DUPOUX, EMMANUEL; and GOLDBERG, YOAV. 2016. Assessing the ability of LSTMs to learn syntax-sensitive dependencies. *Transactions of the Association for Computational Linguistics* 4.521–35.
LINZEN, TAL, and LEONARD, BRIAN. 2018. Distinct patterns of syntactic agreement errors in recurrent networks and humans. *Proceedings of the 40th Annual Conference of the Cognitive Science Society*, Cognitive Science Society, 692–7.